\documentclass[letterpaper, 10 pt, conference]{ieeeconf}  

\IEEEoverridecommandlockouts                              

\usepackage{graphicx}
\usepackage[table]{xcolor}
\usepackage{multirow}
\usepackage{amssymb}
\usepackage{cuted}
\usepackage{mathtools}
\usepackage{array}
\usepackage{tcolorbox}
\let\labelindent\relax
\usepackage{enumitem}
\usepackage{listings}
\usepackage{algorithm}
\usepackage{algpseudocode}
\usepackage{xcolor}
\usepackage[utf8]{inputenc}
\usepackage{wrapfig}
\makeatletter
\let\NAT@parse\undefined
\makeatother
\usepackage{times}
\usepackage{microtype}
\usepackage{amsmath}
\usepackage{booktabs}
\usepackage{makecell}
\usepackage{tabularx}
\setlist[itemize]{leftmargin=1.2em,itemsep=0.15em,topsep=0.2em}
\setlist[enumerate]{leftmargin=1.2em,itemsep=0.15em,topsep=0.2em}
\usepackage{pifont}
\newcommand{\xmark}{\ding{55}}
\definecolor{lightblue}{RGB}{235,244,255}

\usepackage[
  font=footnotesize,
  labelfont=bf,
  textfont=normalfont,
  labelsep=colon,
  justification=raggedright,
  singlelinecheck=true
]{caption}

\usepackage[nocompress,nospace]{cite}
\usepackage{hyperref}

\newcommand{\ours}{P2P-T}

\title{\LARGE \bf
From Pixel to Poses: Object-centric Tool Manipulation Learning \\
from Human Demonstrations
}

\author{
\authorblockN{
Bangjun Wang$^{1}$, Longyan Wu$^{2}$, Yukun Wei$^{1}$, Shenghe Shao$^{1}$, Chaoyi Huang$^{1}$, Wenze Cui$^{1}$, \\ 
Zetong Xu$^{1}$, Hanlin Wu$^{1}$, Long Chen$^{3}$, Yi Ma$^{1}$, Hongyang Li$^{1,2}$
}
\smallskip
\authorblockA{
$^{1}$The University of Hong Kong
~$^{2}$Shanghai Innovation Institute
~$^{3}$Xiaomi EV
}
\smallskip
\authorblockN{\href{https://github.com/OpenDriveLab/P2P-T}{\small\texttt{\textcolor{orange}{https://github.com/OpenDriveLab/P2P-T}}}}
}

\begin{document}

\maketitle
\thispagestyle{empty}
\pagestyle{empty}

\begingroup
\setlength{\stripsep}{2pt}

\begin{strip}
    \vspace{-0.2em}
    \includegraphics[width=\textwidth]{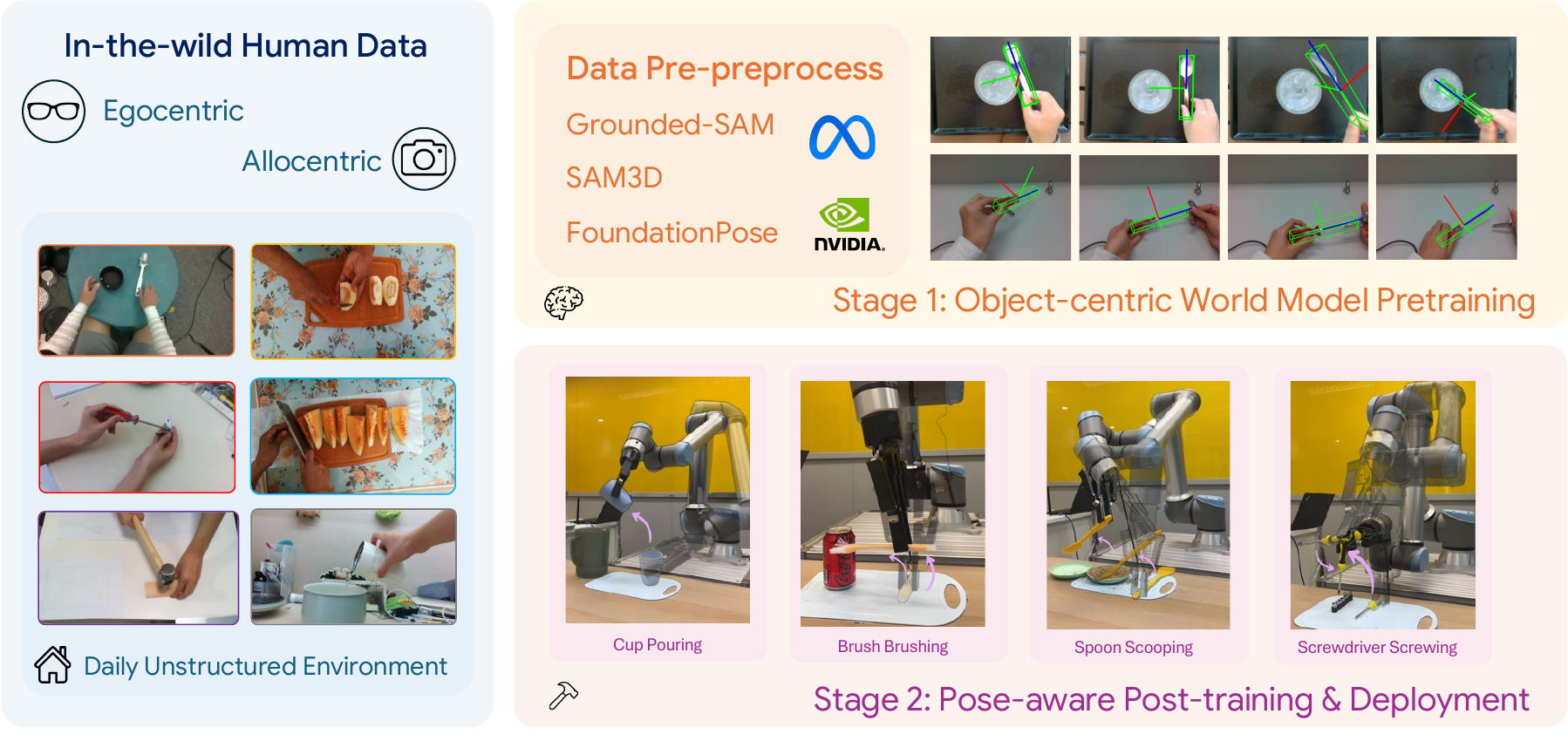}
    \captionof{figure}{\textbf{Overview of the \ours{} framework.} \ours{} leverages in-the-wild human demonstrations (both egocentric and allocentric) to efficiently teach robots complex tool manipulation. Raw video data is first processed through an automated pipeline utilizing foundation models (Grounded-SAM~\cite{ren2024grounded}, SAM3D~\cite{chen2025sam}, and FoundationPose~\cite{wen2024foundationpose}). Stage 1 focuses on pose-aware world model pretraining to extract stable, object-centric pose priors from daily unstructured environments. Stage 2 integrates these learned priors into a low-level execution policy for robot fine-tuning and deployment, successfully bridging the human-robot morphological gap without requiring costly aligned data.}
    \label{fig:teaser_figure}
\end{strip}

\begin{abstract}
    Scaling up robotic manipulation is primarily bottlenecked by the scarcity of real-world robot data. While recent approaches
    leverage human video demonstrations to mitigate this shortage, they remain computationally expensive and still rely on paired human-robot data for domain alignment. 
    Although current state-of-the-arts 
    excel at long-horizon tasks, they struggle with the delicate and precise control required for complex tool manipulation. To overcome these limitations, we introduce \ours{},  
    from \underline{P}ixel to \underline{P}oses for \underline{T}ool Manipulation,
    a data-efficient, object-centric framework that learns tool use directly from human demonstrations. \ours{} bridges the cognitive and physical execution gap through a two-stage approach. 
    First, pretraining an object-centric world model to extract stable pose priors; Second, integrating these priors into an efficient, pose-aware low-level policy. By utilizing a robust automated data processing pipeline powered by modern foundation models, \ours{} completely bypasses the need for human-robot aligned data. This reduces overall training overhead drastically. With minimal per-task fine-tuning, our framework achieves a 73\% improvement over the previous state of the art in execution performance on complex, real-world tool manipulation tasks that currently remain out of reach for standard large-scale pretrained models.
\end{abstract}

\section{INTRODUCTION}

The scarcity of in-domain, real-world robot data has long been the primary bottleneck preventing the field of robotics from scaling up. Collecting paired teleoperation data is notoriously expensive and labor-intensive. To mitigate this shortage, recent approaches such as EgoScale~\cite{zheng2026egoscalescalingdexterousmanipulation} and EgoVLA~\cite{yang2025egovlalearningvisionlanguageactionmodels} have begun to investigate whether models can be directly trained on massive datasets of human videos, or if such data can be integrated into the pretraining recipe. While leveraging human demonstrations presents a highly promising and scalable direction, existing methods are constrained by several practical limitations.

First, these approaches do not fundamentally eliminate the reliance on human-robot aligned data. They still require an additional mid-training or post-training stage using paired datasets to bridge the morphological gap between human hands and robotic end-effectors. Second, the computational cost of extracting motion priors directly from raw human video is exceptionally high. Executing a full-stage training pipeline often demands immense computing resources, sometimes on the scale of hundreds of GB200 GPUs. Finally, these methods suffer from a distinctly limited task scope. Models trained directly on human data are largely restricted to simple pick-and-place actions. Similarly, while current state-of-the-art pre-trained models like 
$\pi_{0}$~\cite{black2026pi0visionlanguageactionflowmodel} and GEN-0~\cite{generalist2025gen0} are highly capable of executing long-horizon planning tasks, they still fall short when confronted with complex tool manipulation. Such tasks inherently demand far more delicate and precise physical control than current generalist models can reliably output.

The importance of mastering tool manipulation cannot be overstated. In dynamic deployments like household environments, it is entirely impractical for robots to frequently change their hardware end-effectors based on the specific task at hand. Therefore, the necessity for a real-world household agent to master the use of tools specifically designed for human hands is self-evident. Confining robots to simple tasks fails to fully exploit their inherent agility and artificially restricts their operational scope. Historically, previous works such as FUNCTO~\cite{tang2025functo} and SimToolReal~\cite{kedia2026simtoolreal} have attempted to solve complex tool-use tasks primarily through rigid rule-based systems or reinforcement learning methods, which often struggle to generalize across various real-world scenarios.

To overcome these challenges, we delve into the possibility of simultaneously instilling both the high-level cognitive capabilities and the low-level executive precision required for tool-use tasks through a highly efficient framework. As shown in Figure.~\ref{fig:teaser_figure}, we propose \textbf{\ours{}}, an efficient, object-centric framework for learning tool manipulation directly from human demonstrations. \textbf{\ours{}} is architected across two distinct stages. In the first stage, rather than attempting to derive dense motion priors directly from unstructured human data, we focus on object-centric world model pretraining. We refine our training objectives to extract object-centric pose priors, which provide a more stable and abstract representation of the task environment. In the second stage, we conduct an efficient, pose-aware policy post-training. During this phase, we seamlessly integrate the pretrained object-centric world model with a low-level execution module. By leveraging the structured pose priors learned during the pretraining phase, we significantly enhance the robot's physical execution capabilities.

In summary, our proposed methodology provides direct solutions to the traditional bottlenecks of robot learning through the following key contributions:
\begin{itemize}
    \item \textbf{Automated Data Pipeline:} Capitalizing on modern foundation models (Grounded-SAM~\cite{ren2024grounded}, SAM3D~\cite{chen2025sam}, and FoundationPose~\cite{wen2024foundationpose}), we construct a robust data processing pipeline that drastically streamlines complex curation procedures and generates clean, low-noise training data.
    \item \textbf{Elimination of Aligned Data:} Our approach successfully bypasses the need for human-robot paired data. By utilizing an object-centric world model design, we completely eliminate the intermediate training stages typically used to map pretrained representations to a robot's physical action space, drastically reducing overall training overhead.
    \item \textbf{Scalable Two-Stage Framework:} We introduce a data-efficient object-centric world model alongside a corresponding pose-aware low-level execution policy. With minimal per-task fine-tuning, \ours{} achieves strong execution performance on complex tool manipulation tasks that currently remain beyond the capabilities of standard, large-scale pretrained models.
\end{itemize}



\section{Related Work}
\label{sec:related-work}

\paragraph{Learning from Human.}
Scaling robot manipulation requires supervision beyond the limited amount of real-world robot data that can be collected through teleoperation. Recent generalist robot policies and vision-language-action(VLA) models have shown that large-scale robot datasets can provide strong policy initializations for downstream manipulation tasks~\cite{team2024octo, kim2024openvla, black2026pi0visionlanguageactionflowmodel, liu2025rdt, intelligence2025pi05visionlanguageactionmodelopenworld}. However, these methods still rely heavily on robot embodiment-specific action data, which remains expensive to collect and often limits the diversity of tasks, scenes, and physical interactions available during training. To overcome this bottleneck, a growing line of work studies how to leverage human videos as a scalable source of behavioral supervision. LAPA learns latent actions from videos without requiring ground-truth robot action labels~\cite{ye2025latent}, while Dreamitate fine-tunes a video generative model on human demonstrations and uses generated execution videos to guide real-world robot control~\cite{liang2024dreamitate}. EgoVLA~\cite{yang2025egovlalearningvisionlanguageactionmodels} and EgoScale~\cite{zheng2026egoscalescalingdexterousmanipulation} further demonstrate the promise of large-scale egocentric human videos for VLA pretraining and dexterous manipulation transfer. Despite their scalability, these approaches usually require either human-to-robot retargeting, paired alignment data, or additional robot fine-tuning to bridge the morphological gap between human hands and robot end-effectors. In contrast, \ours{} avoids directly modeling human hand motion. Instead, it extracts embodiment-independent object pose trajectories from human demonstrations and learns reusable tool-use priors in the object coordinate space.

\paragraph{Tool Manipulation.}
Tool manipulation is substantially more challenging than standard pick-and-place manipulation because it requires reasoning about functional affordances, contact-rich interaction, precise tool orientation, and often forceful or long-horizon execution. Recent works have explored structured representations to improve generalization in tool-use tasks. FUNCTO introduces a function-centric one-shot imitation learning framework that extracts 3D functional keypoints from a human demonstration and establishes correspondences between tools with different geometries~\cite{tang2025functo}. SimToolReal studies zero-shot dexterous tool manipulation by procedurally generating diverse tool-like objects in simulation and training a single goal-conditioned RL policy to manipulate tools toward target poses~\cite{kedia2026simtoolreal}. SPOT further shows that SE(3) object pose trajectories can serve as an effective intermediate representation for object-centric imitation learning, decoupling task constraints from robot-specific actions and enabling learning from action-less human videos~\cite{hsu2025spot}. Although FUNCTO and SPOT are closely related to our focus, neither method is open-sourced, preventing us from including them in our experimental comparisons under the same evaluation protocol. These methods nonetheless highlight the importance of structured object-level representations for tool use. Our work shares this object-centric motivation but differs in both its learning source and policy integration: \ours{} learns pose priors directly from in-the-wild human demonstrations through an automated visual processing pipeline and injects these priors into a low-level, pose-aware action policy for precise execution.

\paragraph{World Model.}
World models aim to learn compact predictive representations of environment dynamics, which can support planning, policy learning, and sample-efficient control~\cite{lingbot-va2026,yang2026rise}. In robotic manipulation, object-centric world models are particularly attractive because manipulation tasks are often defined by how task-relevant objects move and interact. Recent video-based policy learning methods suggest that generative or predictive models can provide useful intermediate plans for downstream control~\cite{liang2024dreamitate}. However, many world-model-based approaches either operate in latent visual spaces or require additional mechanisms to translate predicted dynamics into precise robot actions. \ours{} instead builds a world model over explicit 6D tool poses. This design provides a structured and physically meaningful prediction target that is more directly transferable across embodiments. Moreover, recent foundation models for open-vocabulary segmentation, 3D reconstruction, and 6D pose estimation make it increasingly feasible to extract such structured supervision from raw videos at scale~\cite{ren2024grounded, chen2025sam, wen2024foundationpose}. By combining automated pose extraction with object-centric prediction, \ours{} learns a compact pose prior from human demonstrations and uses it to guide a downstream diffusion-based robot policy.
	

\section{Methods}
\label{sec:methods}

In this section, we detail \textbf{\ours{}}, an efficient two-stage framework for learning complex tool manipulation directly from human demonstrations. Instead of forcing a model to learn morphology-dependent human hand motions from raw videos, our methodology relies on a core insight: while human and robot hands differ fundamentally, the spatial trajectory required to manipulate a specific tool remains consistent. Therefore, \textbf{\ours{}} isolates the task-relevant tool to model its spatial dynamics independently of the embodiment. We first describe an automated pipeline that extracts 6D pose trajectories from human videos. Next, we detail Stage 1, which pretrains an embodiment-independent world model to predict future tool states. Finally, Stage 2 integrates these learned pose priors into a low-level policy to guide precise real-world execution.

\subsection{Human Data Collection and Pre-process}

To construct our Stage 1 pre-training corpus, we recruited human volunteers over a two-month period to capture in-the-wild video demonstrations of diverse tool-use tasks. This self-curated dataset, combined with the existing TACO~\cite{liu2024taco} dataset, comprises our complete pre-training mixture. Detailed statistics are available in Section.~\ref{sec:data-stage1}

While previous approaches are often bottlenecked by brittle, heuristic-based pre-processing methods, we leverage recent advances in foundation models to introduce a streamlined and highly robust pipeline. This approach not only yields high-quality data labels but also significantly reduces annotation costs, paving the way for future scalability.

\subsubsection{Data Distribution of Stage 1}
\label{sec:data-stage1}

The pretraining corpus for the Stage-1 object-centric world model consists of 3,094 video clips spanning 17 distinct tool categories. As illustrated in the data distribution (Figure~\ref{fig:stage1_data}), the dataset exhibits a natural long-tail variance. The most frequently represented tools are those requiring complex, dynamic, or contact-rich interactions, such as spoons (479 clips) and knives (464 clips). These are closely followed by tools that demand precise orientation and force application, including rollers (377 clips), spatulas (343 clips), hammers (264 clips), and brushes (248 clips). 

This concentration of data on highly functional tools is intentional; it ensures the model receives rich supervision to learn robust 6D pose priors for the intricate $SE(3)$ spatial maneuvers required during downstream task execution. Conversely, less frequently represented objects include simpler or less dynamically manipulated items like plates (16 clips), glue (9 clips), and soap (2 clips). Despite this class imbalance, combining our self-curated in-the-wild videos with the existing TACO dataset provides a highly diverse and comprehensive foundation. This variety enables the world model to effectively capture embodiment-independent spatial dynamics, preparing the low-level policy to generalize across a broad array of household tools.

\begin{figure}[h!]
    \centering
    \includegraphics[width=\linewidth]{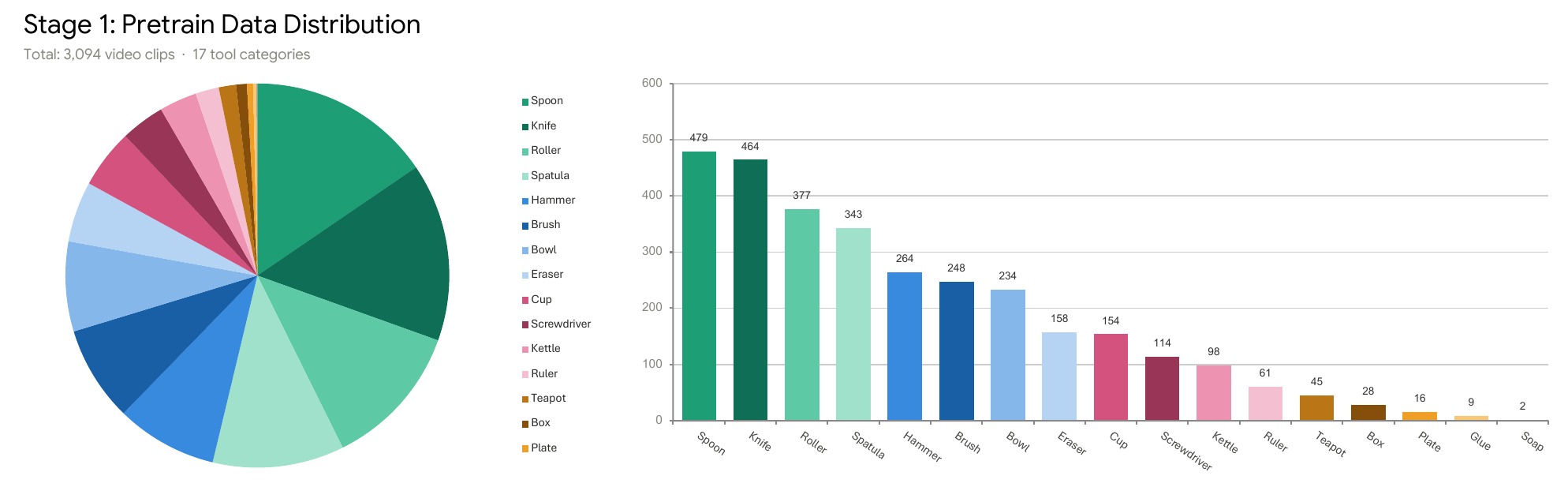}
    \caption{\textbf{Stage 1 Pretrain Data Distribution.} This figure illustrates the composition of the pretraining dataset used for the Stage-1 object-centric world model. The dataset contains a total of 3,094 video clips distributed across 17 different tool categories. The distribution highlights a strong focus on highly interactive and geometrically complex tools, with spoons (479) and knives (464) being the most prevalent. This diverse mixture provides the rich spatial supervision necessary for extracting stable, embodiment-independent pose priors.}
    \label{fig:stage1_data}
\end{figure}

A human verification step was applied during preprocessing to assess the quality of the object tracking results. Clips with inaccurate or unreliable trajectories were discarded, yielding an overall discard rate of 27\%, which we consider acceptable given the in-the-wild nature of the collected videos. The discard rate varies across tool categories according to the complexity of their motion and geometry. For example, cups have a relatively low discard rate of 19\% because their usage typically involves simple and easily trackable motions. In contrast, screwdrivers have a substantially higher discard rate of 37.875\%, as their rotational motion is difficult to track reliably when the tool is approximately symmetric about its $z$-axis. This verification process improves the quality of the retained trajectories and provides more reliable supervision for learning the Stage-1 pose priors.

\subsubsection{Pose Representation}

To accurately capture the spatial dynamics of tool manipulation, we formulate the state of the tool as a 6-Degree-of-Freedom (6DoF) pose within 3D space. Formally, for a given video frame at timestamp $t$, the rigid transformation of the tool relative to the camera coordinate frame is represented by a homogeneous transformation matrix $P_t \in SE(3)$:
\begin{equation}
    P_t = \begin{bmatrix} 
    R_t & T_t \\ 
    \mathbf{0} & 1 
    \end{bmatrix}
\end{equation}
where the translation vector $T_t = [x, y, z]^\top \in \mathbb{R}^3$ denotes the 3D spatial location of the tool's centroid, and the rotation matrix $R_t \in SO(3)$ describes its 3D orientation. 

Although the transformation has six intrinsic degrees of freedom, we retain the matrix representation during training. Specifically, each pose matrix is flattened into a 16-dimensional vector:
\begin{equation}
    p_t = \operatorname{vec}(P_t) \in \mathbb{R}^{16},
\end{equation}
where $\operatorname{vec}(\cdot)$ denotes row-wise vectorization. The vector is reshaped back into a $4 \times 4$ matrix when the predicted pose is decoded or evaluated.
This over-parameterized representation is intentional. Unlike quaternions, which have the sign ambiguity $\mathbf{q}$ and $-\mathbf{q}$ representing the same rotation, and Euler angles, which suffer from angle wrapping and representation singularities, the matrix entries vary continuously with the underlying rotation. Consequently, flattening the homogeneous transformation provides an unambiguous and continuous regression target, avoiding the discontinuous targets that can lead to oscillatory training behavior.

\subsubsection{Pre-processing Pipeline}

\begin{figure}[t!]
    \centering
    \includegraphics[width=\linewidth]{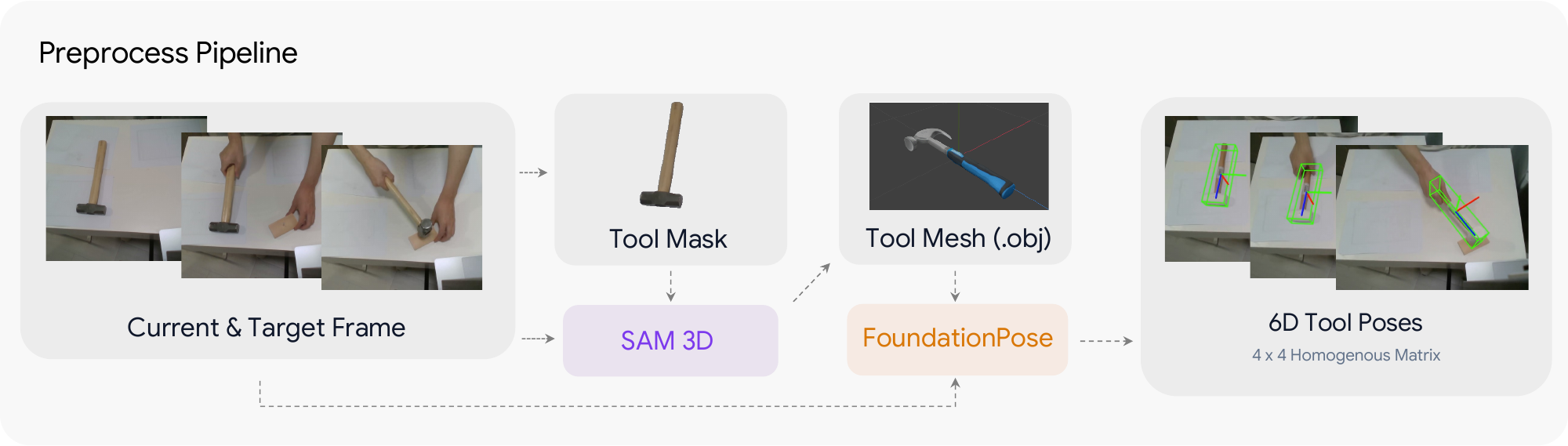}
    \caption{\textbf{Overview of the \ours{} pre-processing pipeline.} We first extract the tool from the background environment to create a 2D segmentation mask. The mask and the raw RGB frames are subsequently fed into SAM 3D~\cite{chen2025sam}, which outputs a high-fidelity, instance-specific 3D tool mesh. Finally, FoundationPose~\cite{wen2024foundationpose} takes the reconstructed 3D mesh and the sequence of raw video frames as inputs to compute and track precise 6D tool poses across the entire demonstration sequence.}
    \label{fig:preprocess}
\end{figure}

Extracting precise 6D tool poses from in-the-wild videos is challenging due to dynamic backgrounds, occlusions, and absent 3D annotations. To address this, we propose an automated pipeline leveraging foundation models for zero-shot 3D reconstruction and tracking (Figure~\ref{fig:preprocess}). For an input sequence $V = \{I_1, \dots, I_N\}$, where $I_t$ is an RGB frame, our pipeline proceeds in three stages:

\begin{enumerate}
    \item \textbf{2D Mask Extraction:} We employ Grounded-SAM~\cite{ren2024grounded} to isolate the target tool and generate a binary segmentation mask $S_t$.
    \item \textbf{3D Mesh Reconstruction:} Feeding the multi-view RGB frames and masks into SAM 3D~\cite{chen2025sam}, we lift the 2D visual boundaries into 3D space to output a high-fidelity tool mesh $M = (\mathcal{V}, \mathcal{F})$.
    \item \textbf{6D Pose Tracking:} FoundationPose~\cite{wen2024foundationpose} aligns the geometric and textural priors of the 3D mesh $M$ with the raw sequence $I_t$ to reliably compute precise 6D tool poses $p_t$ across the entire demonstration.
\end{enumerate}

Ultimately, this automated pipeline transforms raw videos into a dataset of paired $(I_t, p_t)$ samples after human validation. By removing human annotators from the loop, we effectively bypass traditional 3D annotation bottlenecks to provide scalable, high-quality supervision.

\begin{figure*}[t!]
    \centering
    \includegraphics[width=\linewidth]{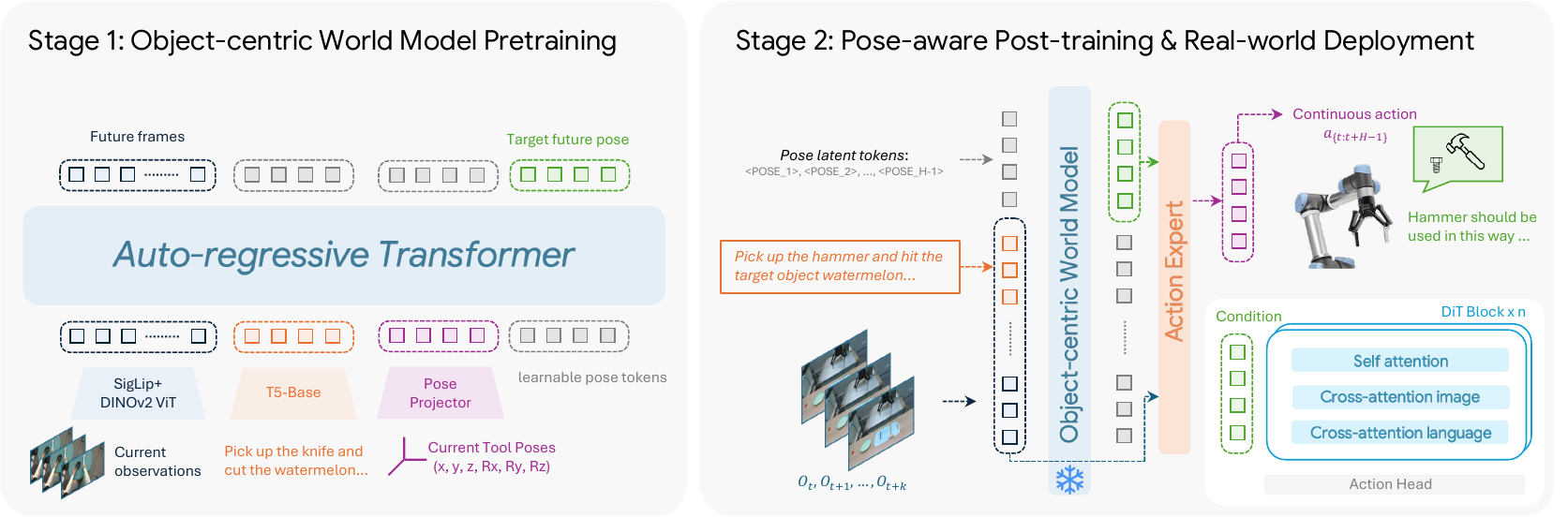}
    \caption{\textbf{Overview of \ours{}'s two-stage framework.} \textbf{Stage 1:} An autoregressive transformer learns an object-centric world model from human demonstrations by fusing observations encoded with SigLIP \cite{zhai2023sigmoid} and DINOv2~\cite{oquab2023dinov2}, instructions encoded with T5~\cite{raffel2020exploring}, and projected tool poses. It learns future pose latents spanning $t+1$ through $t+H-1$, supervised through decoded poses and auxiliary frame predictions. \textbf{Stage 2:} The frozen world model provides future pose latents, mapped through a trainable adaptor to condition an RDT-based diffusion action expert. Together with visual observations, language instructions, and robot proprioception, these latents guide action chunks $a_{t:t+H-1}$ for receding-horizon execution, connecting human-derived object-centric dynamics with robot-specific control without paired human--robot demonstrations.}
    \label{fig:pipeline}
\end{figure*}

\subsection{Stage 1: Object-centric World Model Pretraining}

Stage 1 learns a reusable, object-centric world model from human demonstrations without introducing a robot-specific action space. Its central representation is a sequence of pose latents spanning future time steps $t+1$ through $t+H-1$. These latents capture task-conditioned tool dynamics and are learned through supervision on decoded tool poses, together with auxiliary visual predictions over the same horizon. Since the desired tool trajectory is largely shared across embodiments, this formulation reduces the need for paired human--robot demonstrations.

Each demonstration consists of a sequence of RGB observations $I_t$, a language instruction $c$, and tool poses $p_t \in \mathbb{R}^{16}$, obtained by flattening homogeneous transformation matrices $P_t \in SE(3)$. As illustrated in the left part of Figure~\ref{fig:pipeline}, the world model autoregressively predicts pose and visual latents over a horizon of $H-1$ future steps, conditioned on the current observation, instruction, and tool pose:
\begin{equation}
\left\{
Z_{t+k}^{\mathrm{pose}},
Z_{t+k}^{\mathrm{vis}}
\right\}_{k=1}^{H-1}
=
F_{\theta}(I_t,c,p_t).
\end{equation}
Separate prediction heads decode the latents at each future step into a tool pose and a visual observation:
\begin{equation}
\hat{p}_{t+k}
=
D_{\mathrm{pose}}\left(Z_{t+k}^{\mathrm{pose}}\right),
\qquad
\hat{I}_{t+k}
=
D_{\mathrm{frame}}\left(Z_{t+k}^{\mathrm{vis}}\right),
\end{equation}
$k=1,\ldots,H-1$, The decoded predictions provide supervision across the future trajectory, encouraging the pose latents to capture temporally coherent tool dynamics. The learned pose latents subsequently guide the pose-aware low-level policy in Stage 2.

The world model adopts a transformer-based multimodal fusion architecture. Visual observations are encoded using frozen DINOv2~\cite{oquab2023dinov2} and SigLIP~\cite{zhai2023sigmoid} encoders, while the language instruction is encoded using a frozen T5~\cite{raffel2020exploring} encoder. Freezing these pretrained backbones reduces computational cost and allows the trainable components to focus on task-conditioned spatial and temporal dynamics.

The visual features, language features, and current tool pose are projected into a shared latent space and augmented with learnable modality embeddings. These tokens are concatenated with learnable fusion tokens and processed by an autoregressive transformer to produce the future pose and visual latents. A pose head decodes the pose latents into tool poses, while an auxiliary frame-prediction head decodes the visual latents into the corresponding future observations.

We jointly supervise the decoded pose and visual predictions across the entire prediction horizon. The pose regression objective is
\begin{equation}
\mathcal{L}_{\mathrm{pose}}
=
\frac{1}{H-1}
\sum_{k=1}^{H-1}
\left\|
\hat{p}_{t+k} - p_{t+k}
\right\|_2^2.
\end{equation}
The auxiliary frame-prediction objective provides additional supervision for learning scene dynamics:
\begin{equation}
\mathcal{L}_{\mathrm{frame}}
=
\frac{1}{H-1}
\sum_{k=1}^{H-1}
\left\|
\hat{I}_{t+k} - I_{t+k}
\right\|_2^2.
\end{equation}
The complete Stage 1 objective is
\begin{equation}
\mathcal{L}_{\mathrm{stage1}}
=
\mathcal{L}_{\mathrm{pose}}
+
\lambda \mathcal{L}_{\mathrm{frame}},
\end{equation}
where $\lambda$ balances the two objectives. We set $\lambda=0.01$ to regularize the learned dynamics through visual prediction while prioritizing pose learning. Through this joint supervision over multiple future steps, Stage 1 learns structured, object-centric pose latents that encode future tool trajectories and support policy learning in Stage 2.

\begin{figure*}[t!]
    \centering
    \includegraphics[width=\linewidth]{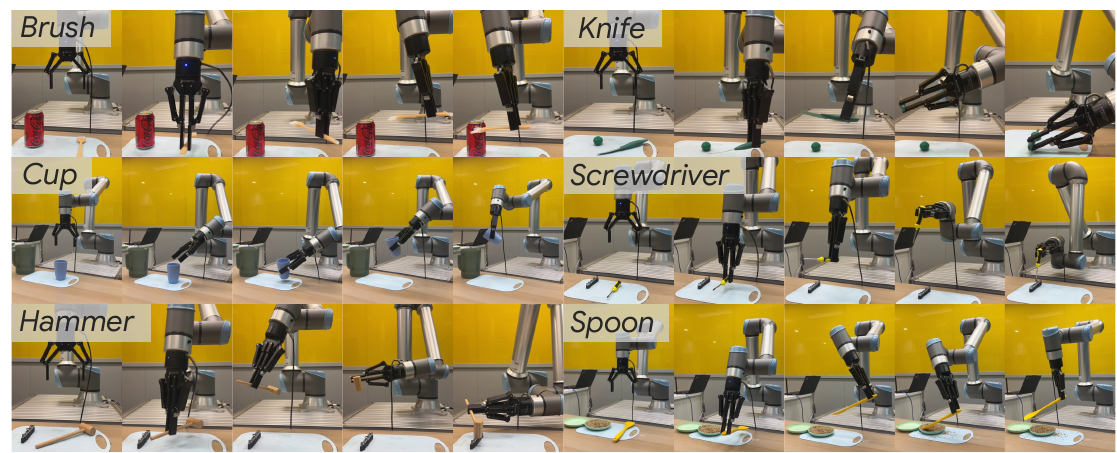}
    \caption{\textbf{Real-world evaluation tasks.} We evaluate on six complex tool-use scenarios that demand delicate physical control and continuous spatial reasoning, moving far beyond standard pick-and-place actions.
    }
    \label{fig:tasks}
\end{figure*}

\subsection{Stage 2: Pose-aware Post-training}

Stage 2 translates the object-centric pose latents learned in Stage 1 into executable robot actions by post-training a low-level action expert initialized from a pretrained RDT~\cite{liu2025rdt} backbone. The action expert conditions on a sequence of future pose latents, allowing it to use the predicted tool trajectory as a structured task-space prior.

Given the current RGB observation $I_t^{c}$ from a designated camera view, the language instruction $c$, and the current tool pose $p_t$ when available, the frozen Stage-1 model produces pose latents spanning time steps $t+1$ through $t+H-1$:
\begin{equation}
\left\{
Z_{t+k}^{\mathrm{pose}}
\right\}_{k=1}^{H-1}
=
F_{\phi}^{\mathrm{pose}}(I_t^{c},c,p_t),
\end{equation}
where $F_{\phi}^{\mathrm{pose}}$ denotes the pose-latent output of the pretrained world model. A shared trainable MLP projects each pose latent token into the RDT embedding space:
\begin{equation}
U_{t+k}^{\mathrm{pose}}
=
\mathrm{MLP}_{p}\left(Z_{t+k}^{\mathrm{pose}}\right),
\qquad k=1,\ldots,H-1.
\end{equation}
The projected tokens are concatenated in temporal order to form
\begin{equation}
U_t^{\mathrm{pose}}
=
\operatorname{Concat}\left(
U_{t+1}^{\mathrm{pose}},\ldots,U_{t+H-1}^{\mathrm{pose}}
\right),
\end{equation}
and inserted before the robot's proprioceptive state token $s_t$ in the RDT input sequence. The action expert thus operates directly on the future pose latents, without requiring decoded pose matrices as policy inputs.

The policy is trained as a conditional diffusion model to predict an action chunk $a_{t:t+H-1}$, conditioned on the pose latent sequence, the proprioceptive state $s_t$, the multi-view RGB history $\mathcal{I}_t$, and the language instruction $c$. The Stage-1 model and the policy's T5~\cite{raffel2020exploring} and SigLIP~\cite{zhai2023sigmoid} encoders remain frozen, while the RDT action expert and pose adaptor are optimized.

During training, Gaussian noise $\epsilon \sim \mathcal{N}(0,\mathbf{I})$ is added to the demonstrated action chunk using a DDPM scheduler~\cite{ho2020denoising}:
\begin{equation}
\tilde{a}^{n}
=
\sqrt{\bar{\alpha}_{n}}\,a_{t:t+H-1}
+
\sqrt{1-\bar{\alpha}_{n}}\,\epsilon,
\end{equation}
where $n$ denotes the diffusion step and $\bar{\alpha}_{n}$ is the cumulative noise-schedule coefficient. We use sample prediction, training the action expert to recover the clean action chunk directly:
\begin{equation}
\mathcal{L}_{\mathrm{act}}
=
\mathbb{E}_{\mathcal{D},n,\epsilon}
\left[
\left\|
a_{\theta}\left(
\tilde{a}^{n},
U_t^{\mathrm{pose}},
s_t,
\mathcal{I}_t,
c,
n
\right)
-
a_{t:t+H-1}
\right\|_2^2
\right],
\end{equation}
where $\mathcal{D}$ denotes the robot demonstration dataset. Conditioning on pose latents across the prediction horizon guides action generation with both the spatial structure and temporal progression of the intended tool trajectory.

During inference, the frozen Stage-1 model first generates the future pose latent sequence. The action expert then conditions on its projected tokens to sample an action chunk using a fast DPM-Solver~\cite{lu2022dpm}. Actions are executed in a receding-horizon manner, with the pose latents and action chunk updated as new observations become available. This architecture connects embodiment-independent tool dynamics with robot-specific control through a learned latent interface.

\begin{table*}[t!]
\centering
\footnotesize
\caption{\textbf{Real-world Policy Evaluation.} We report the success rate for each task, together with the average success rate and inference rate. Each model is fine-tuned with 100 demonstrations and evaluated over 50 trials per task. We report the average results, with the best success rates highlighted in \textbf{bold}. Inference rate is measured in Hz, with higher values being better.}
\label{tab:policy}
\begin{tabular}{l|cccccc|c|c}
\toprule
Method & Hammer & Cup & Brush & Screwdriver & Knife & Spoon &
Average ($\uparrow$) & Execution Rate (Hz) ($\uparrow$) \\
\midrule

\rowcolor{gray!20}
\multicolumn{1}{l|}{\textit{\textbf{Vision-Language-Action Models}}} &
\multicolumn{6}{c|}{} & \multicolumn{1}{c|}{} & \multicolumn{1}{c}{} \\

DP2-DINOv2~\cite{chi2023diffusionpolicy}
& 0.09 & 0.12 & 0.03 & 0.02 & 0.09 & 0.04 & 0.07 & 30 \\
Octo~\cite{team2024octo}
& 0.21 & 0.27 & 0.27 & 0.17 & 0.16 & 0.26 & 0.22 & 15 \\
OpenVLA-OFT~\cite{kim2025fine}
& 0.09 & 0.27 & 0.13 & 0.17 & 0.24 & 0.12 & 0.17 & 25 \\
$\pi_{0.5}$~\cite{intelligence2025pi05visionlanguageactionmodelopenworld}
& 0.24 & 0.49 & 0.27 & 0.35 & 0.37 & 0.24 & 0.33 & 30 \\

\midrule

\rowcolor{gray!20}
\multicolumn{1}{l|}{\textit{\textbf{World Action Models}}} &
\multicolumn{6}{c|}{} & \multicolumn{1}{c|}{} & \multicolumn{1}{c}{} \\

FastWAM~\cite{yuan2026fastwam}
& 0.00 & 0.08 & 0.00 & 0.04 & 0.02 & 0.00 & 0.02 & 10 \\
Lingbot-VA~\cite{lingbot-va2026}
& 0.12 & 0.38 & 0.04 & 0.36 & 0.22 & 0.14 & 0.21 & 8 \\

\midrule
\rowcolor{lightblue}
\ours{} (Ours)
& \textbf{0.59} & \textbf{0.60} & \textbf{0.55} & \textbf{0.54}
& \textbf{0.57} & \textbf{0.58} & \textbf{0.57} & 30 \\

\bottomrule
\end{tabular}
\end{table*}

\section{Experiments}
\label{sec:experiments}

The primary goal of our evaluation is to assess whether \textbf{\ours{}} can effectively bridge the morphological gap and enable precise tool manipulation without relying on paired human-robot data. Unlike conventional evaluations restricted to simple pick-and-place maneuvers, we benchmark our framework on a curated suite of complex tool-use tasks. These tasks require not only a high-level cognitive understanding of functional affordances but also delicate, long-horizon execution involving intricate SE(3) spatial movements.

Our experiments are structured to answer the following core questions:
\begin{enumerate}
    \item How does \textbf{\ours{}} compare against state-of-the-art vision-language-action baselines in complex, real-world tool manipulation tasks? (See Sec.~\ref{sec:main_results})
    \item Does the object-centric world model learned in Stage 1 provide a robust and transferable pose prior for the downstream low-level execution policy? (See Sec.~\ref{sec:ablations})
    \item How does scaling the model size affect the performance of our two-stage framework on complex tool manipulation tasks? (See Sec.~\ref{sec:scaling})
\end{enumerate}

\subsection{Tasks}

To systematically assess our framework's capacity for precise, pose-aware physical execution, we designed a suite of six real-world tasks (Figure~\ref{fig:tasks}) on a UR10e robot with a Robotiq 2F-140 gripper and two D-435 RealSense cameras. Each task challenges a distinct facet of complex spatial manipulation: striking a target with a \textbf{hammer} using dynamic, high-velocity impacts; picking up a \textbf{screwdriver} and rotate it to match with the screw; pouring from a \textbf{cup} while maintaining orientation control; slicing with a \textbf{knife} using precise downward force; sweeping with a \textbf{brush} through sustained surface contact; and scooping with a \textbf{spoon} via coordinated pitch adjustments. During each trial, tools are randomly placed on the table with various initial poses. Models are required to re-orient them to their specific working poses and execute the operation.

By evaluating across these diverse tool categories, we explicitly test the framework's capacity to translate human-designed tool affordances into actionable, robust robot execution, moving far beyond the primitive constraints of standard pick-and-place benchmarks.

\subsection{Implementation Details}
We pretrain the Stage-1 object-centric world model on parsed TACO~\cite{liu2024taco} trajectories and human demonstrations. It processes RGB frames and FoundationPose-estimated~\cite{wen2024foundationpose} 6D tool poses via frozen DINOv2~\cite{oquab2023dinov2}, SigLIP~\cite{zhai2023sigmoid}, and T5-base~\cite{raffel2020exploring} encoders, predicting next tool poses via a transformer. Stage 2 initializes the low-level execution policy from RDT~\cite{liu2025rdt}, conditioned on Stage-1 pose priors. The policy fuses multi-view images, proprioception, and language to predict 64-step action chunks via diffusion.

Octo, OpenVLA-OFT, and $\pi_{0.5}$ use official defaults with adapted interfaces and action horizons of 4/8/50; Octo and $\pi_{0.5}$ use 10 denoising/integration steps. LingBot-VA fine-tunes its ${\sim}5.1$B transformer for 40k steps/task on 4 GPUs (BF16, FSDP, checkpointing, batch size 1/rank, AdamW lr $10^{-5}$, weight decay 0.1). Videos use $256\times320$ resolution at 10 fps (8 actions/latent frame); inference uses 5/10 video/action denoising steps, guidance 5/1, and 16-action chunks at 8 Hz. FastWAM fine-tunes Wan2.2-TI2V-5B video/action DiTs and proprioceptive encoder for 80k steps/task (frozen VAE/text encoders) using BF16, AdamW (batch size 16, weight decay 0.01, peak lr $10^{-4}$, decay $10^{-5}\to 10^{-6}$). Inputs use $224\times224$ images and min--max-normalized states/actions. Inference runs 10 denoising steps for 32-action predictions, executing 5 before replanning at 10 Hz.

\begin{table}[t]
\vspace{-1em}
\centering
\caption{\textbf{Ablations on Framework Design.} }
\begin{tabular}{c|c|cc|c}
\toprule
\makecell{Object-centric\\Pretraining} &
\makecell{Pose-aware\\Post-training} &
Hammer & Cup & Average \\
\midrule
\checkmark & \xmark & 0.12 & 0.14 & 0.13\\
\xmark & \checkmark & 0.32 & 0.36 & 0.34 \\
\midrule
\checkmark & \checkmark & \textbf{0.59} & \textbf{0.60} & \textbf{0.60} \\
\bottomrule
\end{tabular}
\vspace{-0.5em}
\label{tab:ablations}
\end{table}

\subsection{Result Analysis}
\label{sec:main_results}

To evaluate real-world tool manipulation, we compare \textbf{\ours{}} against state-of-the-art vision-language-action baselines (Table~\ref{tab:policy}). Our framework achieves an average success rate of 0.57, substantially outperforming the strongest baseline, $\pi_{0.5}$~\cite{intelligence2025pi05visionlanguageactionmodelopenworld} (0.33), showing an average improvement of 73\%.

Trajectory analysis reveals that while generalist vision-language-action models like $\pi_{0.5}$ exhibit strong initial grasping, they lack explicit knowledge of functional geometry. Consequently, they consistently fail during the interaction phase, struggling with the intricate SE(3) spatial maneuvers required to correctly orient and translate tools to their functional working positions (\textit{e.g.}, aligning a hammer head).

Meanwhile, although world action models (WAMs), such as FastWAM~\cite{yuan2026fastwam} and Lingbot-VA~\cite{lingbot-va2026}, have demonstrated promising generalization on pick-and-place and assembly tasks, they remain inadequate for tool-use scenarios, which require substantially more complex and dexterous manipulation. Given the same number of demonstrations, these models are also more difficult to fine-tune for novel tool-use tasks, as they must learn both the task-specific interaction dynamics and the precise, dexterous motions required to operate the tools, thus resulting in suboptimal performance(0.02 and 0.21 respectively). FastWAM achieves a 0\% success rate on certain tasks, primarily because 100 demonstrations are insufficient for full task adaptation, which is a limitation consistent with our prior experience.Their lower inference rates also reduce responsiveness: FastWAM and Lingbot-VA operate at 10\,Hz and 8\,Hz, respectively, compared with 30\,Hz for \textbf{\ours{}}. 

Conversely, while \textbf{\ours{}} occasionally shows minor grasping instability due to its smaller pretraining scale, this is overwhelmingly compensated for by the robust Stage-1 pose priors. These structured priors provide superior spatial awareness, enabling the policy to precisely reason over the tool's 6D geometry, maintain continuous working orientations, and successfully execute complex tasks.

\begin{table}[t]
\centering
\caption{\textbf{Scaling Verification.} Performance comparison among the \ours{}-Small, \ours{}-Base, and \ours{}-Large variants. ``\#Param.'' denotes the number of parameters.}
\label{tab:scaling}
\begin{tabular}{l|c|cc|c}
\toprule
Model Variant & \#Param. & Hammer & Cup & Average \\
\midrule
\ours{}-Small & 1.189B & 0.34 & 0.48 & 0.41 \\
\ours{}-Base & 1.921B & 0.59 & 0.60 & 0.60 \\
\ours{}-Large & 2.832B & 0.64 & 0.68 & 0.66\\
\bottomrule
\end{tabular}
\vspace{-1em}
\end{table}

\subsection{Ablations}
\label{sec:ablations}
To evaluate the necessity and transferability of the Stage-1 pose priors, we ablate our framework's components (Table~\ref{tab:ablations}).

First, attaching a standard DP2-DINOv2~\cite{chi2023diffusionpolicy} policy to the Stage-1 world model results in poor success rates (0.12 on Hammer and 0.14 on Cup). This suggests that, without a tailored pose-aware architecture, standard policies cannot fully exploit the world model’s geometric priors for complex SE(3) maneuvers.

Second, a fine-tuned RDT~\cite{liu2025rdt} conditioned on the current pose estimated by FoundationPose outperforms the DP2 baseline, achieving 0.32 on Hammer and 0.36 on Cup, but still falls short of satisfactory performance. This suggests that implicit 2D visual features and current-pose information alone are insufficient for fine-grained tool alignment. Pose prior guidance is essential for models to learn those specific tool protocols.

In contrast, our two-stage framework synergizes both components, leaping to 0.59 on Hammer and 0.60 on Cup. This confirms that the Stage-1 world model provides a robust pose prior that, when ingested by the Stage-2 policy, effectively bridges the gap between high-level functional understanding and low-level execution.

\subsection{Scaling Potential}
\label{sec:scaling}

To evaluate the effect of model scaling on performance, we compare three variants of our framework: \ours{}-Small (1.19B parameters), \ours{}-Base (1.92B parameters), and \ours{}-Large (2.83B parameters).

As shown in Table~\ref{tab:scaling}, performance improves consistently with model capacity: \ours{}-Small reaches an average success rate of 0.41, \ours{}-Base reaches 0.60, and \ours{}-Large reaches 0.66. Scaling allows the architecture to absorb more complex spatial distributions. Qualitatively, the larger variants exhibit enhanced generalization to geometric variations (\textit{e.g.}, differing handle lengths or weights) and dynamic disturbances. While the smaller models occasionally require behavioral corrections under extreme variance, the scaled policies yield more robust, smoother action trajectories. This trend indicates that our two-stage framework possesses significant potential to generalize across a broader array of un-modeled household tools as capacity increases.

\section{Limitations}
\label{sec:limitations}
We present \textbf{\ours{}}, an efficient two-stage framework for tool manipulation that bypasses the need for paired human-robot data. By extracting 6D pose trajectories from human demonstrations, we pretrain an object-centric world model to capture transferable spatial dynamics. Integrating these priors into a pose-aware diffusion policy effectively bridges the human-robot morphological gap without costly domain alignment. 

\textbf{Limitation.} Despite its strong performance, labor and compute constraints currently limit our evaluation to a single robot embodiment, precluding the use of multi-fingered dexterous hands. Furthermore, verifying the framework's scaling behavior under massive parameter regimes remains unexplored. Expanding to diverse embodiments, dexterous hardware, and larger-scale training represent critical directions for future work.



\bibliographystyle{IEEEtran}
\bibliography{IEEEabrv, refs}  

\appendices

\end{document}